\documentclass[letterpaper, 10 pt, conference]{ieeeconf}  % Comment this line out if you need a4paper
\IEEEoverridecommandlockouts                              % This command is only needed if
\usepackage{graphics}    % for pdf, bitmapped graphics files
\usepackage{times}       % assumes new font selection scheme installed
\usepackage{amsmath}     % assumes amsmath package installed
\usepackage{amssymb}     % assumes amsmath package installed
\usepackage{graphicx}
\usepackage[separate-uncertainty=true]{siunitx}%
\usepackage[caption=false, font=footnotesize]{subfig}
\usepackage{tabularx}
\usepackage{pgfplots}
\pgfplotsset{width=10cm,compat=1.9}%% Aligns the last page but causes errors on some machines (such as OSX), so don't use it for now.
\usepackage[font=small]{caption}

\def\tabref#1{Tab.~\ref{#1}}
\def\eqref#1{Eq.~(\ref{#1})}

\newcommand\etal{\emph{et al. }}

\usepackage{tabularx}
\newcolumntype{L}[1]{>{\raggedright\arraybackslash}m{#1}}
\newcolumntype{C}[1]{>{\centering\arraybackslash}m{#1}}
\newcolumntype{R}[1]{>{\raggedleft\arraybackslash}m{#1}}
\usepackage{booktabs}
\usepackage{authblk}
\usepackage{url}
\usepackage{caption} % Add caption package
\usepackage{csquotes}

\title{\LARGE \bf RHINO-VR Experience: Teaching Mobile Robotics Concepts in an Interactive Museum Exhibit}

\author{Erik Schlachhoff* \hspace*{2em} Nils Dengler* \hspace*{2em} Leif Van Holland \hspace*{2em} Patrick Stotko\\Jorge de Heuvel \hspace*{2em} Reinhard Klein \hspace*{2em} Maren Bennewitz% <-this % stops a space
	\thanks{* These authors contributed equally to this work.}
 \thanks{All authors are with the University of Bonn, Germany. Nils Dengler, Maren Bennewitz, and Reinhard Klein are additionally with the Lamarr Institute for Machine Learning and Artificial Intelligence, Bonn, Germany. This work has partially been funded by the Deutsche Forschungsgemeinschaft (DFG, German Research Foundation) under the grant numbers BE 4420/2-2 and KL 1142/11-2 (FOR 2535 Anticipating Human Behavior) and by the Federal Ministry of Education and Research within the project BNTrAinee (funding code 16DHBK1022).}}

\begin{document}
\maketitle
\thispagestyle{empty} 
\pagestyle{empty}

%%%%%%%%%%%%%%%%%%%%%%%%%%%%%%%%%%%%%%%%%%%%%%%%%%%%%%%%%%%%%%%%%%%%%%%%%%%%%%%%
\begin{abstract} 
  % ´Personalized robot behavior is the next level.
  % WHY is it relevant?
  % WHICH PROBLEM do we address?
  % HOW is our approach special, WHAT are we actually doing, and WHAT IS NEW
  %% IMPLEMENTATION, EVALUATION, WHAT FOLLOWS 
In 1997, the very first tour guide robot RHINO was deployed in a museum in Germany. 
With the ability to navigate autonomously through the environment, the robot gave tours to over 2,000 visitors. 
Today, RHINO itself has become an exhibit and is no longer operational
%Today, RHINO has become an exhibit itself and is no longer functional.
In this paper, we present RHINO-VR, an interactive museum exhibit using virtual reality (VR) that allows museum visitors to experience the historical robot RHINO in operation in a virtual museum. 
RHINO-VR, unlike static exhibits, enables users to familiarize themselves with basic mobile robotics concepts without the fear of damaging the exhibit.
%Compared to static exhibits, RHINO-VR allows users to overcome the fear of damaging the exhibit while familiarizing them with the basic concepts of mobile robotics. 
In the virtual environment, the user is able to interact with RHINO in VR by pointing to a location to which the robot should navigate and observing the corresponding actions of the robot. 
To include other visitors who cannot use the VR, we provide an external observation view to make RHINO visible to them. 
We evaluated our system by measuring the frame rate of the VR simulation, comparing the generated virtual 3D models with the originals, and conducting a user study. 
The user study showed that RHINO-VR improved the visitors' understanding of the robot's functionality and that they would recommend experiencing the VR exhibit to others.
%Additionally, the evaluation revealed that a constant frame rate and an accurate 3D model of the environment highly benefit the experience of the visitors.

\end{abstract}

%%%%%%%%%%%%%%%%%%%%%%%%%%%%%%%%%%%%%%%%%%%%%%%%%%%%%%%%%%%%%%%%%%%%%%%%%%%%%%%%
\section{Introduction}
\label{sec:intro}

%Nowadays, mobile robots are a crucial part of our daily life.
Due to their versatility, mobile robots have become an integral part of many aspects of the modern world and are used in a wide range of applications. 
However, most people are not familiar with the underlying mechanisms and tend to be afraid of unfamiliar and unknown technology. 
This fear of autonomous robots and artificial intelligence (FARAI)~\cite{liang_fear_2017} may explain, among others, people's prejudices towards human-robot interaction (HRI). 
Therefore, in order to reduce FARAI, people need to increase their knowledge about the topic.
A typical place where people of all ages can voluntarily educate themselves is a museum, where they can learn through observation and activity \cite{brajcic_learning_2013}.

%However, most people are not familiar with the underlying technology, which increses the gap . 
%Due to the increasing use of those robots in our everyday life,  to familiarize onself with the robots and the basic principals behind them. 
%
%However, for many poeple that do not work with robotic systems, many basic principles of modern systems are not known increasing the gap to familiarize with them. 
%In the past there were many efforts made to close this gap and strengthn the coexistence between human and robots.
One of the earliest attempts to include robots in our everyday life was the interactive tour guide robot RHINO \cite{burgard_rhino_1999}, which was developed to guide visitors through the exhibits in a museum. 
In 1997, over a period of six days, RHINO was deployed at the Deutsches Museum Bonn in Germany, providing tours to more than 2,000 visitors.
%In 1997, RHINO was deployed in the \enquote{Deutsches Museum Bonn} in Germany for a period of six days and gave tours to over 2,000 visitors. 
In addition, people from all over the world were able to watch and control RHINO’s operation through an interactive web interface. 
Equipped with four different sensor systems and the latest developments in the field of robotics and artificial intelligence, RHINO was able to freely navigate through the museum while avoiding collisions with visitors and other obstacles. 
The success of RHINO motivated researchers to develop an improved version called MINERVA~\cite{thrun_minerva_1999}, which was deployed for two weeks in the Smithsonian’s National Museum of American History one year later. 
Therefore, RHINO and MINERVA are two of the first known robots interacting autonomously with humans within a real-world environment for a longer period of time. 
They created the foundation for several decades of research in the field of mobile robotics. 
Today, after more than 25 years, RHINO is not functional anymore and is a museum exhibit itself, located at the \enquote{Deutsches Museum Bonn}.

 \begin{figure} 
	\centering
	\subfloat{%
		\includegraphics[width=0.85\linewidth]{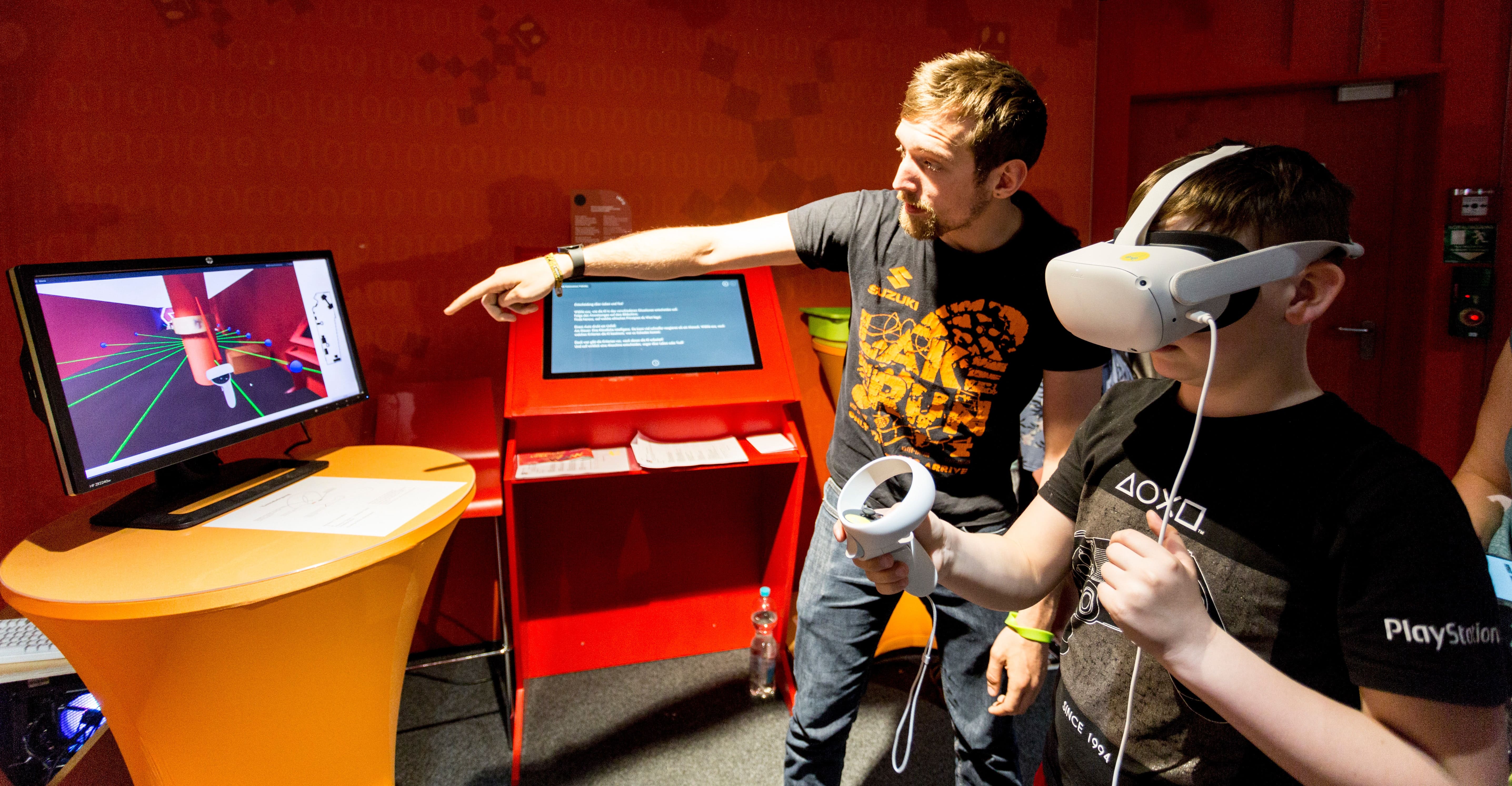}}
	\\
	\subfloat{%
		\includegraphics[width=0.85\linewidth]{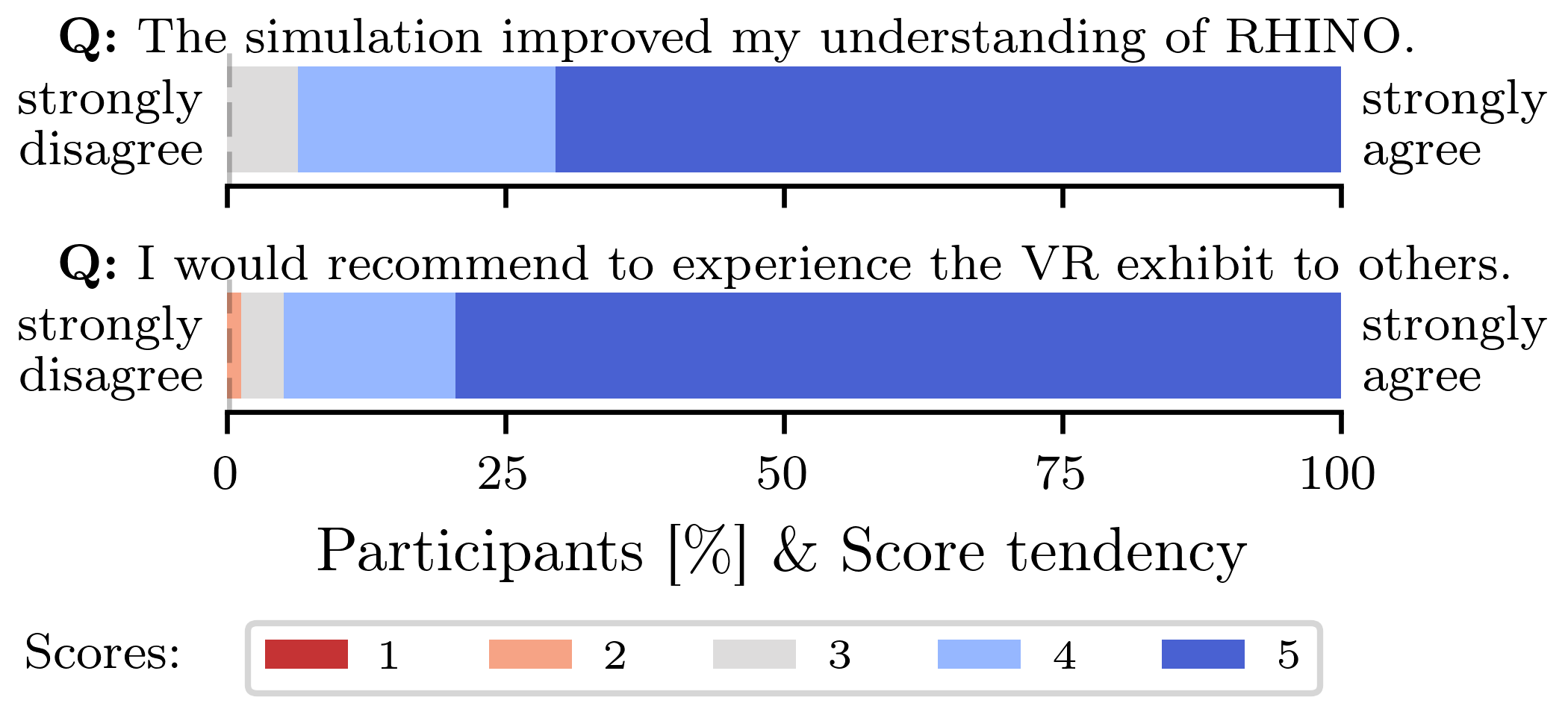}}
	\caption{
		\textbf{Top:} We present an interactive virtual reality (VR) museum exhibit that explains the basic concepts of mobile robotics to visitors. The user can indicate a navigation goal to the robot by pointing at it with a handheld controller. Information about perceived sensor data is shown to the user within the VR headset, but also on an external screen for other interested visitors.
		\textbf{Bottom:} User study survey results on the importance of the interactive exhibit to increase the knowledge about the presented paradigms.
		\label{fig:motivation}}
\end{figure}

%Nowadays, mobile robots are a crucial part of our daily life. 
%A large subset of these operates closely with humans and interacts with them. 
%However, most people are not familiar with the underlying technology. 
%People tend to be afraid of unfamiliar and unknown things. 
%Carleton [2] named it the \enquote{Fear of the Unknown} (FOTU) and has revealed, that this is a fundamental fear and a key component of anxiety. 
%Liang and Lee [3] took the concept of fear and combined it with the field of autonomous robotics. 
%They have introduced the fear of autonomous robots and artificial intelligence (FARAI) and used it to explain people’s prejudices of human-robot interaction (HRI). 
%FARAI reflects the likelihood that people expect more negative experiences such that they anticipate being anxious. 
%The authors found that the education has a negative impact on level of fear. 
%People with a bachelor’s degree or higher tend to have a lower level of FARAI compared to people with a college degree or lower.
%Thus, to reduce the fear of autonomous robots, people have to increase their knowledge about this topic.
%A typical place where people can choose to educate themselves voluntary is a museum. 
%Individuals of all ages can visit museums and learn through observation and activity [4]. 

Over the last few years, virtual reality (VR) systems have become increasingly popular in museum environments. 
They allow people to immerse into virtual worlds, e.g., a reconstructed historical place \cite{shehade_vr_museums_2020}, and can be used to bring otherwise lost knowledge back to life \cite{bekele2018survey}. Such exhibits make use of the learning-by-doing approach \cite{reese_learning_2011}. 
Therefore, the goal of this work is to create an interactive museum exhibit that aims to introduce the basic concepts of mobile robotics to visitors, i.e., the ability to perceive the world and to move and navigate through it according to the information perceived, to reduce the FARAI. 
To achieve this, we present RHINO-VR, an interactive museum exhibit using VR that allows museum visitors to experience the historical robot RHINO and its functionalities.
Visitors can control the virtual RHINO, while learning about its sensors, and navigation algorithms.
To reduce the virtual reality gap and enhance the experience of the users, the presented exhibit employs a digital reconstruction of the museum as a 3D scan where the exhibit is deployed.

%% MAIN CONTRIBUTION & WHAT FOLLOWS FROM THAT
In summary, the main contributions of our work are:
\begin{itemize}
\item An interactive VR museum exhibit that explains basic concepts of mobile robotics to visitors and attempting to reduce the FARAI.
\item A digital reconstruction of the historical deployment of the RHINO robot, as well as the digital reconstruction of the museum environment, enhancing visitor engagement and understanding.
\item A user study that shows the benefits of interactive exhibits to make important concepts graspable.
\end{itemize}

%%%%%%%%%%%%%%%%%%%%%%%%%%%%%%%%%%%%%%%%%%%%%%%%%%%%%%%%%%%%%%%%%%%%%%%%%%%%%%%%
\section{Related Work}
\label{sec:related}
\subsection{Virtual Reality Applications}
%\begin{tikzpicture}
%\begin{axis}[
%    ybar,
%    enlargelimits=0.15,
%    legend style={at={(0.5,-0.15)},
%      anchor=north,legend columns=-1},
%    ylabel={Frequency},
%    symbolic x coords={uncomfortable, comfortable},
%    xtick=data,
%    nodes near coords,
%    nodes near coords align={vertical},
%    ]
%\addplot coordinates {(uncomfortable,0.4) (comfortable,0.2)};
%\addplot coordinates {(uncomfortable,0.7) (comfortable,0.3)};
%\addplot coordinates {(uncomfortable,0.6) (comfortable,0.5)};
%\legend{Ours,SC,DWA}
%\end{axis}
%\end{tikzpicture}

Virtual reality has a wide range of different applications, from entertainment and gaming \cite{hartmann_entertainment_2021} through industry and healthcare \cite{javaid_medical_2020} to robotics \cite{perez_industrial_2019}.
Within this domain, VR systems enable personalized robot navigation behavior through demonstrations, as evidenced by de Heuvel \etal \cite{de_Heuvel_2022}. 
They also facilitate safe testing in human-robot interaction (HRI), mitigating risks to both robots and humans from unexpected behaviors \cite{bottega_virtual_2022}.
%In this domain, VR systems offer the possibility to personalize robot navigation behavior through demonstrations, as shown by de Heuvel \etal \cite{de_Heuvel_2022}, and can be used for testing of applications in the field of HRI, without the risk of damaging crucial parts of the robot as well as  minimizing risk of injuries due to unexpected robot behavior \cite{bottega_virtual_2022}.
Furthermore, VR systems offer the possibility of exploration and teleoperation, as presented by Stotko \etal \cite{stotko_slamcast_2019,stotko_exploration_2019}. 
In their system a 3D model of the environment is reconstructed in real time based on RGB-D data allowing users to immersively explore the environment while wearing a VR headset. 

In our work we leverage the potential of VR systems for educational purposes, by considering safety and fear aspects, as in  \cite{bottega_virtual_2022, javaid_medical_2020}, immersively introduce users to the concept of mobile robotics, using methods of \cite{stotko_slamcast_2019}, and letting the users interactively decide where the robot should go next, as shown in \cite{de_Heuvel_2022}.

\subsection{Interactive Museum Exhibits}
Museums play an important role in education, serving as a major source of learning outside schools \cite{komarac2023understanding}.
%Museums play an important role in education as being a major source for learning outside schools \cite{komarac2023understanding}. 
In order to provide information more efficiently, many museums started to offer interactive experiences to visitors in multiple ways such as mobile apps for smartphones \cite{andritsou_momap_2018}, touch-sensitive screens and digital animations \cite{lindemann_collecting_2022}, and virtual and augmented reality~(AR) applications \cite{cristobal_coral_2020,beheshti_wires_2017}. 
%Especially the latter two create many different opportunities for museums to interact with their visitors. 
Virtual and augmented reality, in particular, open numerous opportunities for museums to engage with visitors.

In general, there is shown a great benefit in re-imagining cultural heritage applications in VR in the current literature  \cite{bekele2018survey, tsita2023virtual, leow2021analysing}, highlighting its importance and potential.
For instance, VR allows modelling of virtual environments that visitors can explore, such as an ancient Aztec city \cite{garcia_tenochtitlan_2017}, or a Viking encampment from the 9th century \cite{schofield_viking_2018}. 
However, it deployed on a phone-based VR system with limited computational capabilities, leading to a poorly visual experience.  
The immersive experience provided by VR systems was particularly highlighted by users, which inspired our work.
%Especially the immersion using a VR system was highlighted by many users and motivated our work. 
A conducted user study in \cite{schofield_viking_2018} showed that especially the immersion using a VR system was highlighted by the majority of users and motivated our work.
%The reconstruction of the virtual environment was made by hand based on archaeological and historical evidence.
%Due to the fact, that the virtual scene of the RHINO application represents the actual museum where the exhibit is deployed rather than a historical place, the environment is reconstructed automatically using a scan of the museum.
Our work employs a system with a Meta Oculus Quest~2 as the display device as well as a dedicated high-end computer for rendering which, in turn, has more computational power compared to phone-based VR systems. 
Furthermore, the system includes handheld controllers, enabling enhanced movement and interaction within the virtual environment.
%Additionally, the system is equipped with handheld controllers enabling movement in and more interaction with the virtual environment. 
%The authors of Viking VR evaluated their application using a short questionnaire, where 83\% of the visitors gave positive feedback. 
%Especially, the immersion using a VR system was highlighted by many users. 
%AR applications have similar effects on visitors, as presented in \cite{beheshti_wires_2017}. 
%The implemented exhibit was designed to make electrical circuits more accessible and understandable to museum guests. 
%By conducting a study, the authors showed that AR has a positive effect on learning. 
%Our application is evaluated employing a user study that incorporates a questionnaire, similar to the evaluation of Viking VR, to analyze the educational effect of VR as well as the experience visitors make with this system.

%%%%%%%%%%%%%%%%%%%%%%%%%%%%%%%%%%%%%%%%%%%%%%%%%%%%%%%%%%%%%%%%%%%%%%%%%%%%%%%%
\section{Our Approach}
\label{sec:Approach}
\begin{figure*}[ht!]
	\centering
	\includegraphics[width=0.95\linewidth]{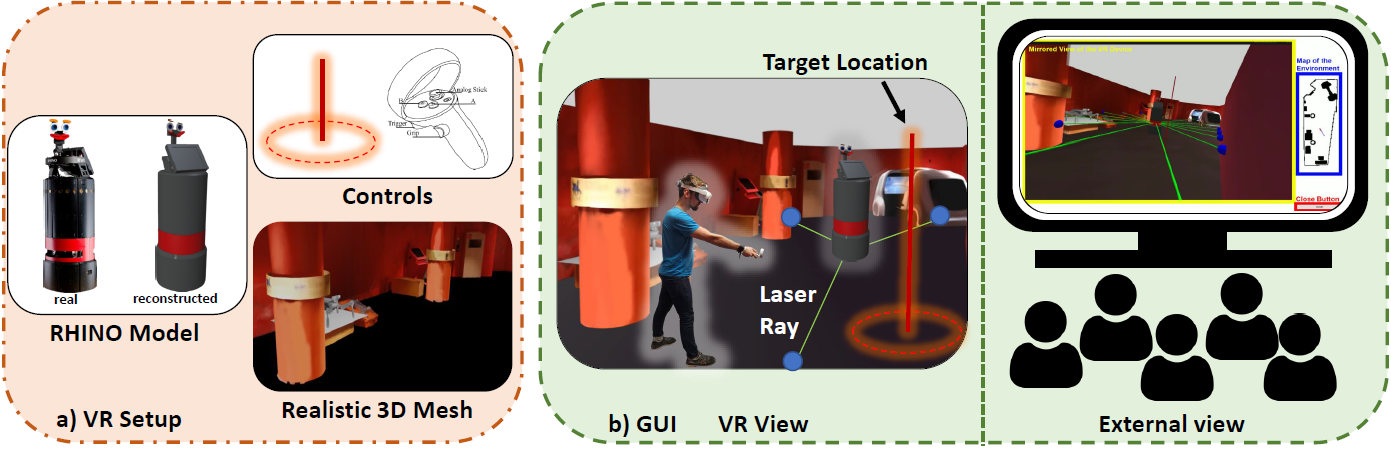}
	\caption{Schematic representation of the used VR-setup and the graphical user interface (GUI). The VR setup consists of a realistic model of the RHINO robot and the museum environment, as well as the controlling unit of the VR. The GUI is separated into the internal VR view and the external view for visitors that are not using the VR headset but are still able to observe the robot and its functionality.}
	\label{fig:architecture}
\end{figure*}

In the following, we first give an overview of our system and introduce our framework in detail afterward.

\subsection{Overview}
In this work, we aim to develop an interactive VR museum exhibition, to teach the foundations of robotics and artificial intelligence behind the mobile robot RHINO. 
A schematic overview of our approach is shown in Fig.~\ref{fig:architecture}. 
The project focuses on creating a virtual version of the museum, accessible through a VR headset. 
Within this digital environment, visitors can explore a fully operational, autonomous replica of the RHINO robot. 
Interaction is facilitated by allowing visitors to direct the robot's movements by pointing to desired locations within the virtual space. 
As the robot navigates, educational content about its functionality is displayed via visuals and text. 
To accommodate multiple viewers, the VR experience is simultaneously broadcasted on an external screen. 
This screen not only shows the robot's activities but also provides detailed insights into its localization process, enabling visitors without the headset to engage with the exhibit.

\subsection{Implementation and Design of the VR Application}
For the VR application we use the iGibson simulation environment \cite{li_gibson_2022} which facilitates compatibility with standard VR headsets.
%through OpenVR\footnote{\url{https://github.com/ValveSoftware/openvr/wiki/API-Documentation}}.
%an API that simplifies interactions with VR devices without requiring detailed hardware knowledge.
The application's graphical user interface (GUI) is designed to provide an immersive experience to museum visitors. 
It includes a red laser pointer for selecting the robot's target within the virtual environment, an informational text box displaying facts about the historical robot RHINO, and a visualization of the robot's sensor rays to illustrate environmental perception. 
The content of the text box, which changes according to the current state of RHINO, is shown in Tab. \ref{tab:text_box}. 
These features combined offer an interactive and educational experience.

Control within the VR environment is achieved through handheld controllers of the VR headset, while
positional movements of the visitor are mirrored in the virtual world by the VR's functionalities. Due to the space constraints in museum settings, larger movement of the user within the virtual environment is also controlled via the analog stick on the right controller. 
Moreover, the earlier mentioned laser pointer originates from the right controller, which allows visitors to point at locations on the ground to indicate navigation goals for the robot.

Overall, the VR application aims at providing an engaging and informative interface for museum visitors, combining intuitive controls and educational elements within a realistically simulated environment.

\subsection{3D Museum Models}
\begin{figure}[t!]
	\centering
    \includegraphics[width=0.96\linewidth]{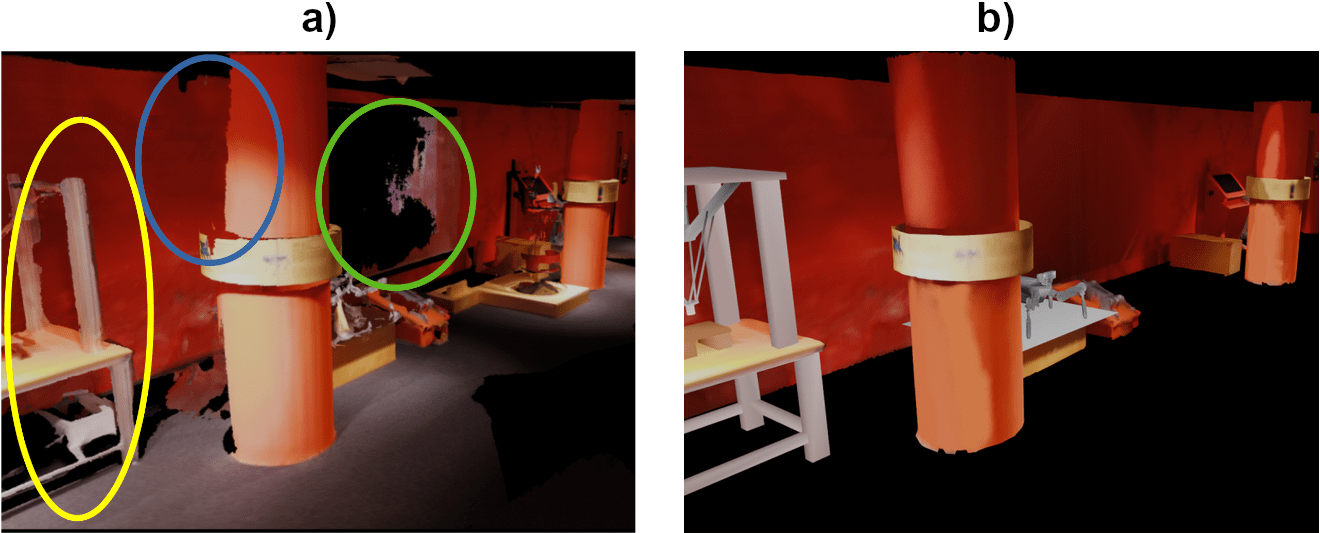}
	\caption{Qualitative comparison of the 3D model before \textbf{a)} and after \textbf{b)} post-processing. Qualitative changes are shown in green, yellow, blue}
	\label{fig:3d_model}
\end{figure}

%\begin{figure}[t!]
%	\centering
%    \subfloat[\label{3da}]{
%	\includegraphics[width=0.48\linewidth]{figures/before_1.png}
%    }
%    \subfloat[\label{3db}]{
%    \includegraphics[width=0.48\linewidth]{figures/after.png}
%    }
%	\caption{\todojdh{use consistent a) b) fig indicators at top of figure, compare Fig. 2+5} Qualitative comparison of the 3D model before (a) and after (b) post-processing. Qualitative changes are shown in green, yellow, blue}
%	\label{fig:3d_model}
%\end{figure}
The environmental model is reconstructed from a sequence of RGB-D images captured using an Azure Kinect camera. 
The reconstruction process utilizes a client, as detailed in \cite{stotko_slamcast_2019}, to fuse the collected data into a coherent 3D model.
Here, the camera is localized using standard SLAM techniques and the RGB-D data are integrated into an implicit truncated signed distance function (TSDF). 
After the environment has been captured, a 3D model in terms of a dense triangle mesh is extracted from the TSDF representation using marching cubes. 
The depth sensor operates on the time of flight (ToF) principle, measuring the time an infrared signal takes to reflect from an object to the sensor.  
This method, while effective, is susceptible to inconsistent raw measurements resulting in missing depth data, particularly when encountering dark or highly reflective surfaces.
To mitigate this issue and enhance the quality of the reconstructed model, we employ Blender, a versatile 3D computer graphics software\footnote{\url{https://www.blender.org}}. 
Blender plays a crucial role in refining the model to fill small holes and to compress the resolution of the model for fast rendering while retaining intricate texture details.

The resulting model, seamlessly integrated into the VR application, represents a significant improvement in realism and detail. 
Fig.~\ref{fig:3d_model} illustrates a section of the captured museum model, showcasing the enhancements achieved through manual post-processing in Blender. 

\begin{table}[t]
\centering
\begin{tabularx}{1.\linewidth}{c X}
\midrule
\vspace{0.1em}a) & In May 1997, the barrel-shaped robot guided museum visitors to selected exhibits and presented them. Now it has been virtually recreated.\\[1em]
\midrule
\vspace{0.1em}b) & You are now in a room with RHINO. Move the right controller in any direction and press the A button to give RHINO a destination \\[1em]
\midrule
\vspace{0.1em}c) & To determine its position in space, the robot first scanned its surroundings with laser and sonar sensors. RHINO compared the result of the scan with a predefined map and estimated its probable position. After each movement, RHINO calculated new possible positions. \\[0.5em]
\midrule
\vspace{0.1em} d) & RHINO used a movement planning algorithm called $A^\ast$ to avoid obstacles. It calculated several paths based on its current position, speed and surroundings. It then chose the shortest path to reach a specific destination - without colliding with obstacles!\\
\midrule
\vspace{0.1em} e) & To avoid people walking around the museum, RHINO constantly updated a map during its journey. An artificial neural network was used for this and trained in advance to predict whether there is an obstacle in its path. \\
\midrule
\vspace{0.1em} f) & RHINO was also equipped with infrared and touch sensors, as well as two cameras. The camera was also used at the time to allow people to view the exhibition and control RHINO remotely from home. \\
\midrule
\end{tabularx}
\captionsetup{type=table} % Explicitly specify this is a table caption
\caption{Interactive snippets presented to visitors in VR, based on the current state of the robot.}
\label{tab:text_box}
\end{table}

\subsection{Transferred Navigation Concepts}
To enhance the visitor experience in a virtual museum environment, our model closely emulates the navigation capabilities of the original RHINO system. 
This section introduces the core functionalities required for autonomous navigation that will be presented to visitors in the proposed VR exhibit, emphasizing the seamless integration of laser-based localization and dynamic path planning mechanisms.

Upon receiving a navigation target from the user, the virtual RHINO system activates its internal state machine and employs global Markov localization \cite{fox1999markov}. 
This is done using 180 unique laser beams that are sampled at 2-degree intervals around the robot. 
The data collected by these sensors is then processed using a method known as the likelihood field model \cite{thrun_probabilistic_2005}, which helps the robot predict its position with high accuracy by comparing the sensor data with a pre-existing map of the environment. 
The laser beams and their endpoints are visualized for the user in VR and for others on an external screen.

For path planning, RHINO uses the information provided by localization to plan its movements. 
It calculates the best path to the target location using the $A^\ast$ path planning algorithm \cite{hart_paths_1968}. 
To avoid obstacles in real time, it uses the Dynamic Window Approach (DWA) \cite{fox_dynamic_1997}, which calculates safe velocities for the robot to move without colliding with any object.
This involves evaluating several possible motions and choosing the one that optimizes the robot's speed, distance from obstacles, and direction to the goal.

Finally, the robot integrates its movement decisions with its internal model of the world, adjusting its beliefs about its location based on the executed movements. 
Unlike the original RHINO, which used real-world velocity data, our virtual robot uses an odometric motion model \cite{thrun_probabilistic_2005}. This model is better suited to the virtual environment because it allows the robot to accurately calculate its movements based on the simulated data available. 
%This step involves complex calculations \todo{what are these complex calculations? that term sounds suspicious.} to ensure that the robot's belief about its position remains as accurate as possible, even as it moves.

In essence, our virtual robot navigates through its environment by continuously collecting data, processing this information to understand its position, planning safe and efficient paths to its destination, and adapting its internal model based on its movements.
This navigation system allowed the robot to effectively guide visitors through the museum in 1996, and is reimplemented in our VR exhibit to teach visitors playfully about the important concepts that are still used in modern robots. The whole process is explained to the viewers within the text boxes shown in Tab. \ref{tab:text_box}.c) to e).

\subsection{External GUI}
\label{sec:external_gui}
\begin{figure}[t!]
	\centering
	\includegraphics[width=\linewidth]{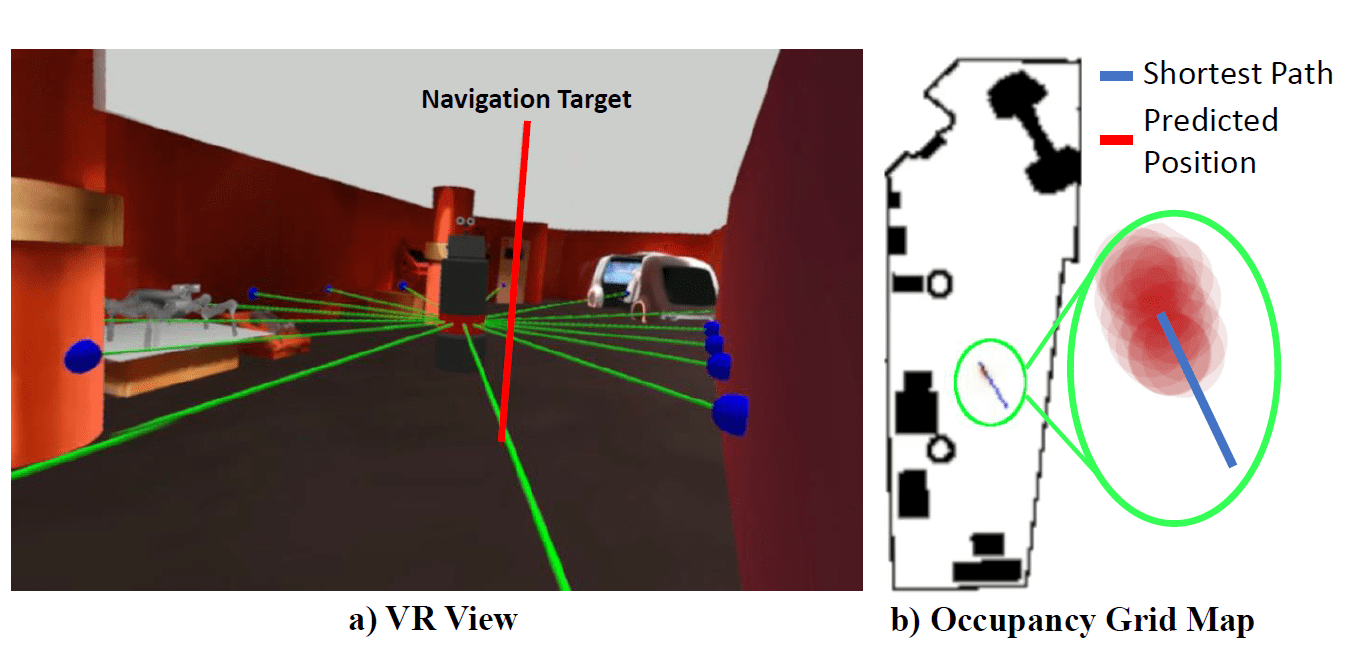}
	\caption{External screen representation. The user can perceive information from the VR view (a) and the map of the environment (b) including the robot's position (marked in green).}
     \label{fig:external_gui}
\end{figure}
The final component of our implementation is an external GUI, designed to provide information to museum visitors not using the VR headset, developed with PyQt5.
%\footnote{\url{https://www.riverbankcomputing.com/software/pyqt/}} 
Our GUI comprises two main elements:

\begin{itemize}
    \item \textbf{Mirrored VR View:} The central feature of the GUI is the mirrored display of the VR headset's view. This is achieved by leveraging the mirroring capability of the SteamVR system. PyQt is used to embed this view within our GUI. (Refer to Fig.~\ref{fig:external_gui}.a). 
    
    %\todo{These parts are hard to identify in the figure as they are not really labeled there.}
    
    \item \textbf{Virtual Environment Map:} The map, displayed on the right side of the GUI, dynamically represents the robot's belief state in shades of red, with more intense red indicating a higher probability. Additionally, the robot's planned path is depicted in blue. (See Fig.~\ref{fig:external_gui}.b).

\end{itemize}
Through these elements, the external GUI not only enhances the experience for non-VR users but also aims at providing an intuitive and informative interface, making the exploration of the virtual museum and the RHINO robot both engaging and educational.

\begin{figure}[t!]
	\centering
	\includegraphics[width=0.95\linewidth]{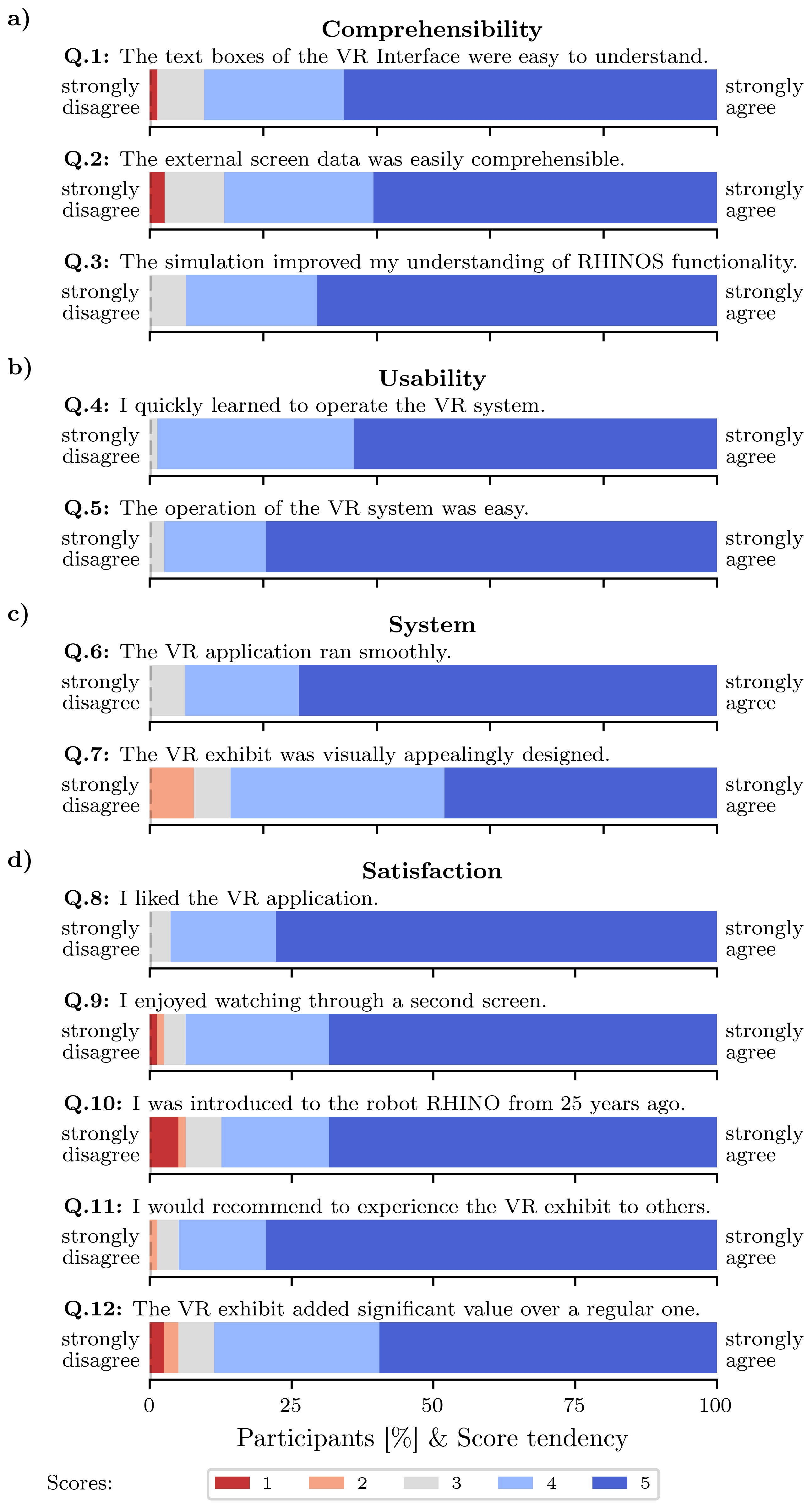}
	\caption{User study survey results. \textbf{a)} The information supplied by the application was predominantly assessed as easily comprehensible. \textbf{b)} All participants perceived the application as user-friendly \textbf{c)} The virtual reality exhibit was acknowledged for its visually appealing design, and participants noted that it operated smoothly. \textbf{d)} Our application was mostly perceived positively, and virtually all visitors would recommend it. 
	\label{fig:survey_all}}
\end{figure}

\section{Experimental Evaluation}
\label{sec:exp}

This section highlights the results of our user study and provides an analysis of the performance of the application as well as the accuracy of the virtual models. All experiments are performed with the Oculus Quest 2 VR headset on a computer with an NVIDIA RTX 3080 graphics card.

\begin{figure*}[t]
\centering
% Row 1
\subfloat[\label{modela}]{
        \includegraphics[width=0.13\linewidth]{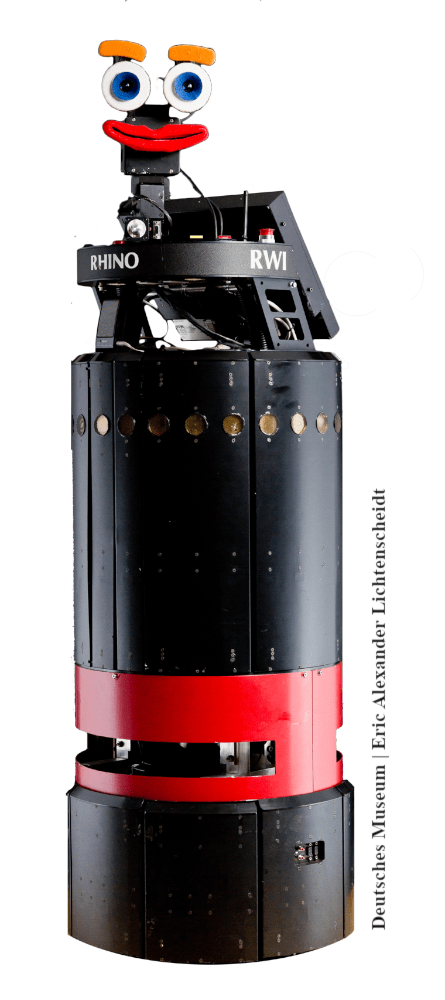}
        \includegraphics[width=0.13\linewidth]{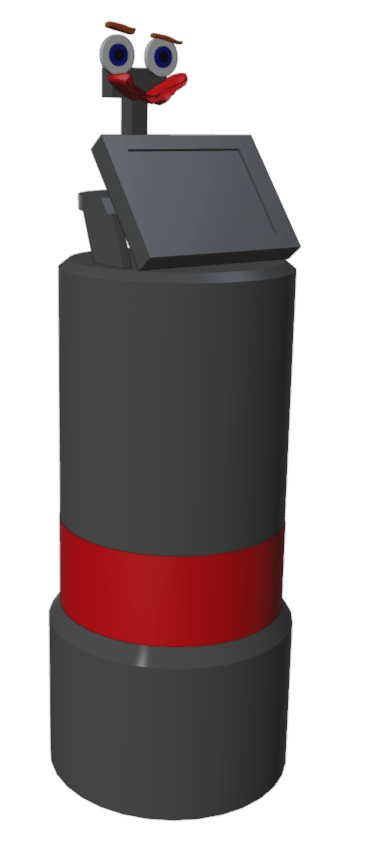}}%
\subfloat[\label{modelb}]{
        \includegraphics[scale=0.5]{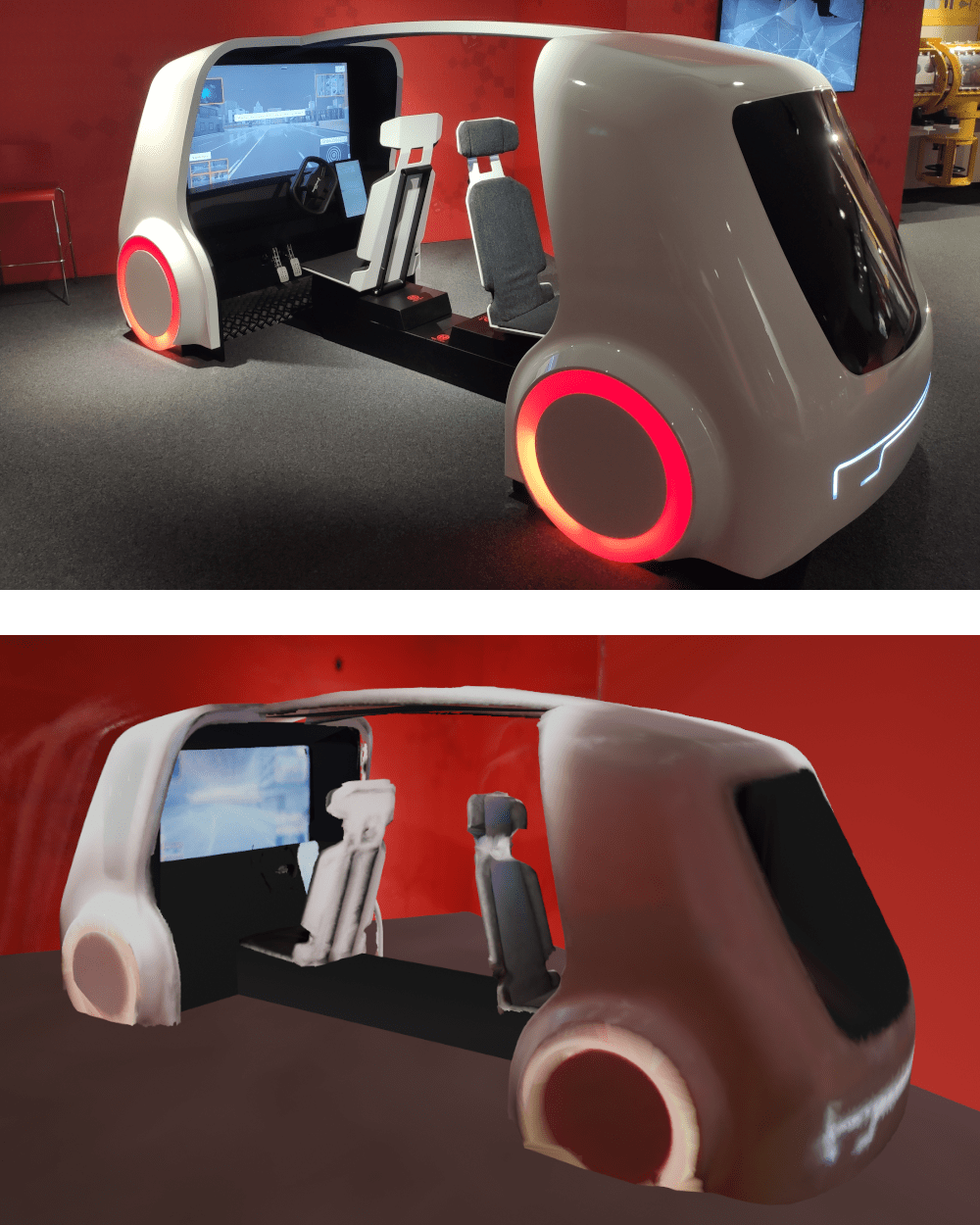}}
\subfloat[\label{modelc}]{
        \includegraphics[scale=0.5]{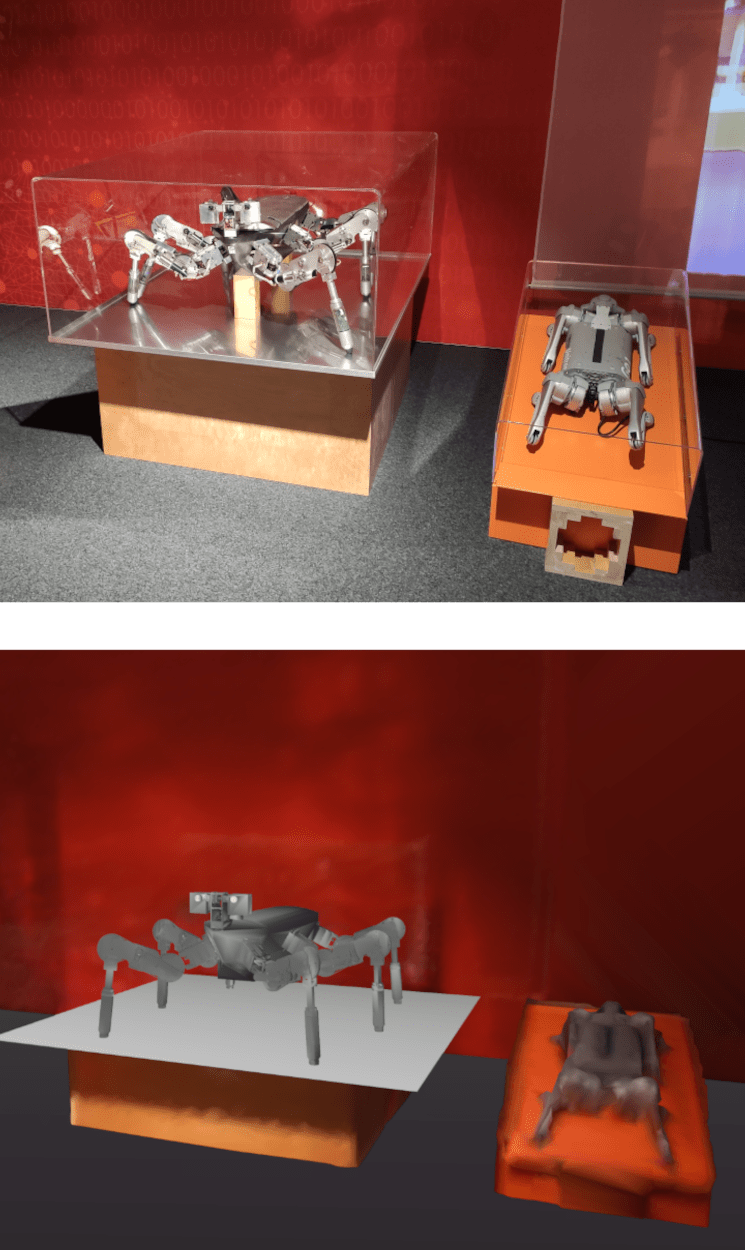}}
\subfloat[\label{modeld}]{
        \includegraphics[scale=0.5]{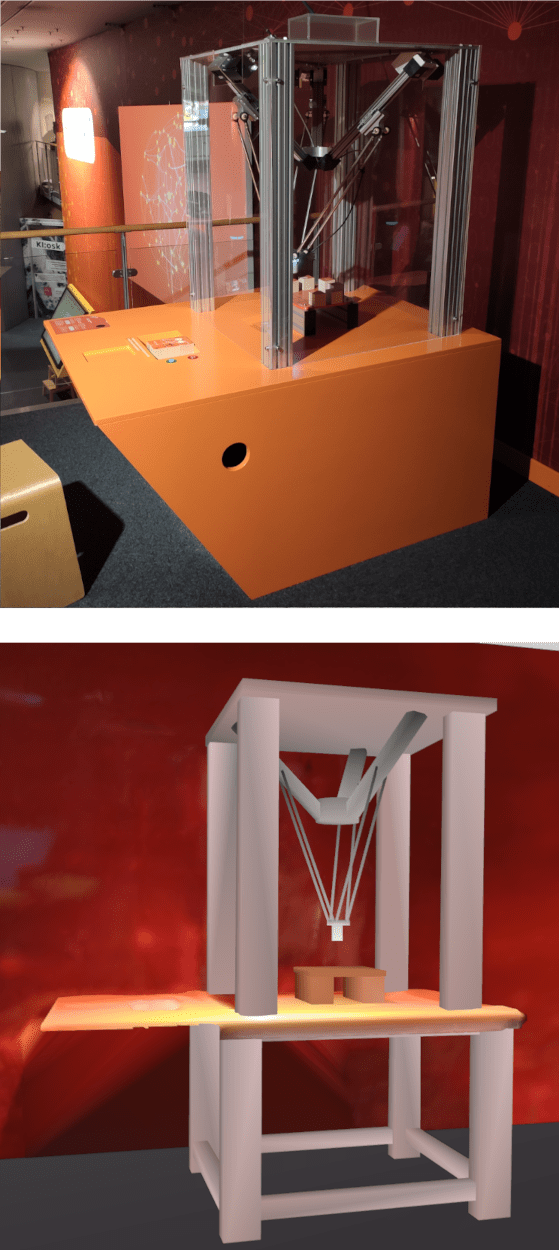}}
\subfloat[\label{modele}]{
        \includegraphics[scale=0.5]{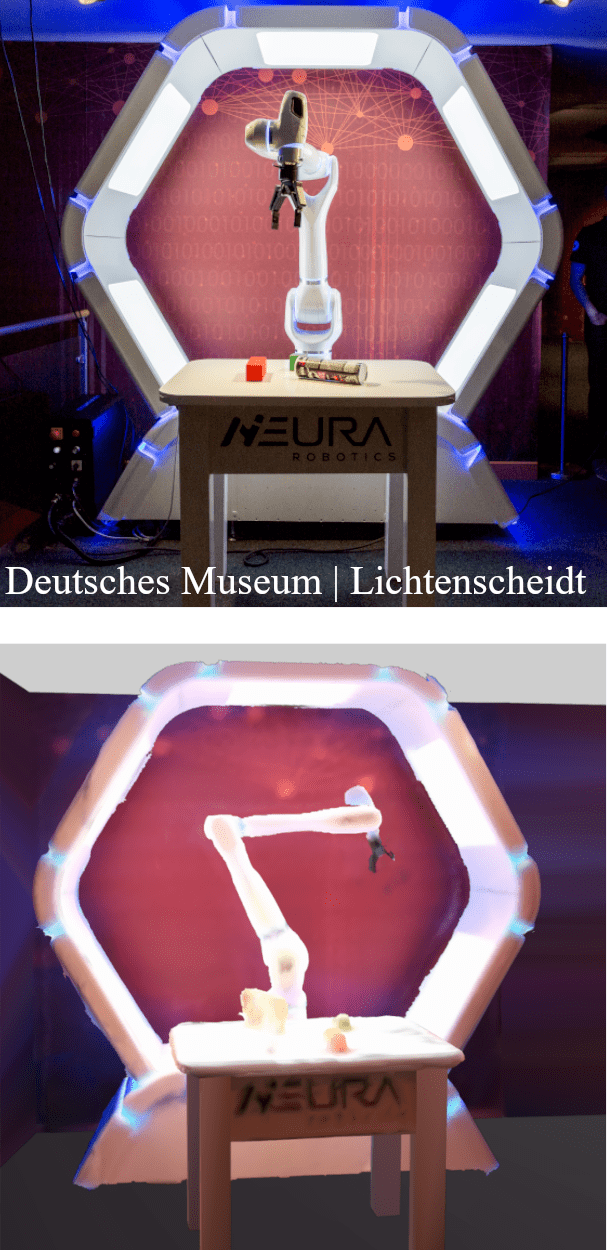}}
\caption{Comparison of physical and virtual models. \textbf{a)} RHINO, \textbf{b)} Driving Simulator, \textbf{c)} Lauron IVc and Unitree Go1, \textbf{d)} Puzzle \enquote{Human against Machine}, and \textbf{e)} Neura Robotics MAiRA. As can be seen, the 3D models precisely reconstructed the real exhibits in the museum with only minor differences.}
\label{fig:models}
\end{figure*}

% USER STUDY
\subsection{User Study}
In order to gain insights into the user experience, acceptance, and effectiveness of the implemented application, we conducted a user study over a period of two days at the \enquote{Deutsches Museum Bonn}, the place where the original RHINO is currently located at, with a total of 85 participants.
We designed the study in a way that all participants either operated the VR system themselves or observed the operating user through the external screen, as described in Sec. \ref{sec:Approach}. 
After experiencing the exhibit, we asked the participants to anonymously complete a questionnaire to evaluate the application. Each participant had the opportunity to ask questions at any time if something was unclear during the questionnaire. 
The design of the questionnaire is inspired by \cite{de_Heuvel_2022} and \cite{tucker20museum} with changes made according to our particular use case, as described in the following.

The evaluation and design of the questionnaire employ a mixed methods approach consisting of quantitatively items, to measure the quality of the implementation regarding \textbf{i)} the comprehensibility, \textbf{ii)} usability, \textbf{iii)} the system, and \textbf{iv)} user satisfaction, and a qualitative, open question item to identify potential areas for improvement which have not been addressed by the closed items. 
Please note, that items related to usability were only answered by individuals who operated the VR system.

We measure the closed items on a 5-point Likert scale $M$ ranging from \enquote{strongly disagree} ${M=1}$ to \enquote{strongly agree} ${M=5}$ with a \enquote{neutral choice} at ${M=3}$. 
Our hypothesis~$H1$ before conducting the user study is that the mean score of all criterions is greater than the midpoint of the Likert scale, meaning that our implementation meets certain quality standards when the individual areas exceed the evaluation criterion of ${M = 4}$. 
The survey questions and results are shown in Fig.~\ref{fig:survey_all}, while \tabref{tab:evaluation} shows the significance test results of one-sample t-test, with all results being significant regarding $H1$.

We now evaluate the study in more detail, where the values in brackets refer to the mean survey scores~(1-5) and their standard deviations.
The assessment of comprehensibility, as illustrated in Fig.~\ref{fig:survey_all}.a) indicates a clear benefit by the provided text boxes \mbox{$(Q1: 4.5 \pm 0.8)$} and the external screen \mbox{$(Q2: 4.4 \pm 0.9)$}.
%Moreover, participants reported a high level of clarity in understanding the instructions of the VR system, reflected by an average rating of \mbox{$(Q3: 4.6 \pm 0.7)$}.
Additionally, all visitor stated, that the application significantly contributed to their understanding of RHINO's functionality \mbox{($Q3: 4.6 \pm 0.6$)}. 
Since a good comprehensibility is important for real understanding of a topic, these significant results are promising in terms of the positive influence of our VR exhibit.

Furthermore, the usability assessment \ref{fig:survey_all}.b) reveals that all participants rated the VR application easy to use \mbox{($ Q5: 4.8 \pm 0.5$)}, indicating an effortless entry into using the application, which makes it accessible to a wide range of people. 
This low barrier of usage is also reflected in the rapid learning curve in operating the VR system \mbox{($Q4: 4.6 \pm 0.5$)}.
However, especially a good system performance is key for a smooth and valuable experience.
As Fig. \ref{fig:survey_all}.c) shows, the majority of the participants are satisfied with the performance as well as the cisual design of the 3D models and reflect this in a high score \mbox{($Q6: 4.7 \pm 0.6; Q7: 4.2 \pm 0.9)$}.
The performance will be evaluated in more detail in Sec. \ref{sec:Performance}.

Finally, we evaluate the satisfaction of the participants when using the VR exhibit. 
The very good and significant score of \mbox{$(Q8: 4.7 \pm 0.5)$}, as shown in Fig. \ref{fig:survey_all}.d), demonstrates the high value of our system and again summarizes the over all good scores of comprehensibility, usability, and system performance. 
With the received positive feedback from the participants \mbox{($Q9: 4.6 \pm 0.7$)}, the incorporation of an external screen for observing the application is a crucial part of the system since it includes people into the exhibit that typically cannot use it due to difficulties, such as motion sickness or age restrictions. 
Since the success of exhibitions can significantly depend a lot on direct recommendations received from related persons like the family or friends, we evaluated a high willingness to recommend our system \mbox{($Q11: 4.7 \pm 0.6$)}, which could relate to the fact that it offered a distinct added value compared to a conventional exhibit \mbox{($Q12: 4.4 \pm 0.9$)}, and is a promising result.

In summary, the user study results in majorly positive feedback from the participants, which expressed that the robot from 25 years ago was effectively introduced to them \mbox{($Q10: 4.4 \pm 1.0$)}.
For this reason, our system was chosen to be a permanent exhibit in the \enquote{Deutsches Museum Bonn}.

As next step, we plan to extend the exhibit to function in augmented reality, enabling the virtual robot to autonomously navigate within the actual museum space.
Addressing the open item of the user study which asked for further suggestions of improvement, participants mentioned the need for various environments as well as expanding the existing environment in size which will be added in one of the next versions of the system.

%%%%%%%%%%%%%%%%%%%%%%%%%%%%%%%%%%%%%%%%%%%%%%%%%%%%%%%%%%%%%%%%%%%%
% TABLES
%%%%%%%%%%%%%%%%%%%%%%%%%%%%%%%%%%%%%%%%%%%%%%%%%%%%%%%%%%%%%%%%%%%%%
\begin{table}[!t]
\centering
\begin{tabular}{C{0.1\linewidth} C{0.15\linewidth} C{0.1\linewidth} C{0.125\linewidth} C{0.125\linewidth}} 
\toprule
 & & & \multicolumn{2}{c}{95\% confidence interval} \\ \cmidrule{4-5}
Q & T & df  & lower & upper \\
 \midrule
 Q.1 & $5.965^{***}$ & 72 &  4.356 & 4.713\\ 
 Q.2 & $4.155^{***}$ & 75 &  4.219 & 4.623\\ 
 Q.3 & $9.397^{***}$ & 77 &  4.505 & 4.777\\ 
 Q.4 & $10.56^{***}$ & 74 &  4.508 & 4.745\\ 
 Q.5 & $14.112^{***}$ & 77 &  4.661 & 4.878\\ 
 Q.6 & $10.224^{***}$ & 79 &  4.544 & 4.806\\ 
 Q.7 & $2.548^{**}$ & 76 &  4.057 & 4.463\\ 
 Q.8 & $12.843^{***}$ & 80 &  4.626 & 4.856\\ 
 Q.9 & $6.953^{***}$ & 78 &  4.416 & 4.749\\ 
 Q.10 & $3.806^{***}$ & 78 &  4.211 & 4.675\\ 
 Q.11 & $10.83^{***}$ & 77 &  4.596 & 4.865\\ 
 Q.12 & $3.943^{***}$ & 78 &  4.201 & 4.610\\ 
 \bottomrule
 \multicolumn{5}{l}{\scriptsize{${}^*p<0.05, {}^{**}p<0.01, {}^{***}p<0.001$}}
\end{tabular}
\captionsetup{type=table} % Explicitly specify this is a table caption
\caption{Results of the one-sample t-tests with test value $M=4$, meaning that our implementation meets certain quality standards when the individual areas exceed the given evaluation criterion.}
\label{tab:evaluation}
\end{table}
%%%%%%%%%%%%%%%%%%%%%%%%%%%%%%%%%%%%%%%%%%%%%%%%%%%%%%%%%%%%%%%%%%%%%

\subsection{Performance}
\label{sec:Performance}
Frame rate plays an important role in the field of VR applications, as it directly affects the user’s experience and sense of immersion. 
A low frame rate not only reduces the overall immersion but can also lead to motion sickness. 
To evaluate our application regarding the performance, we measured the average frame rate over a time period of 10~minutes. 
%\textbf{The industry standard for developing a VR game is 90 FPS.} 
%This value was taken as a reference when evaluating the museum application. 

Our application ran with an average of 36.6\,fps while the worst 1\% \,fps of the application averaged around 34.3\,fps and the worst 0.1\% \,fps averaged around 32 \,fps.
While this performance leads to a smooth user experience, the frame rate is limited by the non-optimized iGibson renderer and can be further increased with better hardware and a stronger simulation engine such as the Unreal Engine or Unity.%, which are possible areas for future work.

\subsection{Virtual Models}

To ensure that the virtual environment closely resembles the museum, both the environmental model and the model of RHINO were evaluated for architectural and visual similarity with their physical counterparts.
Fig \ref{fig:models} illustrates individual exhibits located in the specified museum area alongside their corresponding counterparts in the virtual environment.

It can be observed that the models exhibit very high visual similarities to the originals. 
Individual differences can be attributed to various factors at the time of scanning, such as the absence of the lower cover in Fig.~\ref{fig:models}.d) or the highly reflective surface of the exhibit in Fig.~\ref{fig:models}.c). 

In addition, we performed measurements of the virtual 3D model sizes in comparison to the real-world objects, uncovering an average deviation of 1.03\%. 
A comparable evaluation for the RHINO model highlighted a size discrepancy of 2.8\%. 

In conclusion, the comprehensive evaluation of both environmental and robotic models reveals that the virtual representations closely resemble their physical counterparts, with discernible differences attributed to specific factors during the scanning process.

\section{Discussion and Future Work}
The user study demonstrates the RHINO-VR exhibit's significant impact on the understanding and engagement of the visitors with mobile robotics concepts. Furthermore, the interactive VR exhibit effectively conveys ideas in an accessible and engaging manner. The combination of the museum's reconstructed scene, the VR system, and real-time interaction with the RHINO robot provided an immersive experience, allowing users to understand how mobile robots perceive their environment, avoid obstacles, and plan paths.

\subsubsection{User Experience}
One key goal of RHINO-VR was to enhance visitors' learning experiences by providing an interactive and immersive environment. Positive feedback from the user study confirmed that visitors found the VR system easy to use and highly informative. The external screen, which mirrored the VR experience, was particularly effective in including observers who were not using the VR headset, ensuring a broader audience engagement. The visual and textual explanations of RHINO's functionalities within the accurate 3D reconstructed VR environment helped visitors to understand the robot's operational principles more deeply.

\subsubsection{Challenges and Solutions}
Implementing the VR exhibit involved challenges, especially in ensuring a smooth runtime performance and high-quality visualizations. The average frame rate of 36.6 fps provided a relatively smooth experience, though optimization or hardware upgrades could further improve this. The iGibson simulation environment facilitated realistic interactions, but future versions could use advanced rendering engines like Unreal Engine or Unity for better visual fidelity and performance. Additionally, a permanent connection to a powerful PC is required. We plan to develop a lightweight version that solely runs on the internal VR system for more location-independent scenarios.

\subsubsection{Reduction of FARAI}
Liang~\etal~\cite{liang_fear_2017} suggest that user-friendly applications may reduce FARAI. The user study results in terms of usability (Fig. \ref{fig:survey_all}.b) and satisfaction (Fig. \ref{fig:survey_all}.d) indicate that RHINO-VR provides a better understanding with an easy-to-use nature, supporting the claim by~\cite{liang_fear_2017}. We will further investigate the system's influence on reducing FARAI in more detail in future work.

\subsubsection{Potential for Augmented Reality}
The RHINO-VR exhibit's promising results may inspire future developments such as integrating augmented reality (AR). This could further enhance the exhibit's realism and educational value by allowing the virtual RHINO to navigate through the actual museum space. Expanding the exhibit to include various environments and larger spaces would also address user feedback and improve the educational impact.

%%%%%%%%%%%%%%%%%%%%%%%%%%%%%%%%%%%%%%%%%%%%%%%%%%%%%%%%%%%%%%%%%%%%%%%%%%%%%%%%
%\section{Discussion}
%\label{sec:discussion}
%Addressing the open item of the user study which asked for further suggestions of improvement, participants mentioned the need for various environments as well as expanding the existing environment in size which could be added in later versions.

\section{Conclusion}
\label{sec:conclusion}
We presented an interactive VR museum exhibit introducing museum visitors to mobile robotics concepts through a virtual replica of the historic robot RHINO. 
Utilizing a VR headset, visitors successfully instructed RHINO to navigate autonomously within the virtual museum, accompanied by explanations through visual effects and text. 
%The virtual scene, created using a 3D scan of the museum, faithfully replicated the real environment of the exhibition. 
Created from a 3D scan of the museum, the virtual scene accurately replicated the exhibition's real environment.
The user study indicated that visitors perceived both the usability of the application and the provided information as easily comprehensible, contributing to their understanding of the fundamental concepts of RHINO. 
%Our study underscores the importance of reviving old technologies through VR exhibits, making them accessible and engaging for all, particularly younger generations.
%Our study showed that it is of great importance to not forget about old technologies but to relive them as VR-Exhibit to make it accessible and relatable to everyone, especially the younger generations. 
%Moreover, the study's findings led to the VR application's positive reception and its permanent installation in the museum.
%Furthermore, the study revealed that the VR application garnered positive reception among visitors, leading to its establishment as a permanent exhibit in the museum. 

%%%%%%%%%%%%%%%%%%%%%%%%%%%%%%%%%%%%%%%%%%%%%%%%%%%%%%%%%%%%%%%%%%%%%%%%%%%%%%%%
% Only if applicable
% \section*{Acknowledgments}

\bibliographystyle{IEEEtran}
\bibliography{bibliography}

\begin{thebibliography}{10}
\providecommand{\url}[1]{#1}
\csname url@rmstyle\endcsname
\providecommand{\newblock}{\relax}
\providecommand{\bibinfo}[2]{#2}
\providecommand\BIBentrySTDinterwordspacing{\spaceskip=0pt\relax}
\providecommand\BIBentryALTinterwordstretchfactor{4}
\providecommand\BIBentryALTinterwordspacing{\spaceskip=\fontdimen2\font plus
\BIBentryALTinterwordstretchfactor\fontdimen3\font minus
  \fontdimen4\font\relax}
\providecommand\BIBforeignlanguage[2]{{%
\expandafter\ifx\csname l@#1\endcsname\relax
\typeout{** WARNING: IEEEtran.bst: No hyphenation pattern has been}%
\typeout{** loaded for the language `#1'. Using the pattern for}%
\typeout{** the default language instead.}%
\else
\language=\csname l@#1\endcsname
\fi
#2}}

\bibitem{liang_fear_2017}
Y.~Liang and S.~Lee, ``Fear of autonomous robots and artificial intelligence:
  Evidence from national representative data with probability sampling,''
  \emph{International Journal of Social Robotics}, vol.~9, 2017.

\bibitem{brajcic_learning_2013}
M.~Brajčić, S.~Kovačević, and D.~Kuščević, ``Learning at the museum,''
  \emph{Croatian Journal of Education}, vol.~15, 2013.

\bibitem{burgard_rhino_1999}
W.~Burgard, A.~B. Cremers, D.~Fox, D.~Hähnel, G.~Lakemeyer, D.~Schulz,
  W.~Steiner, and S.~Thrun, ``Experiences with an interactive museum tour-guide
  robot,'' \emph{Artificial Intelligence}, vol. 114, no.~1, 1999.

\bibitem{thrun_minerva_1999}
S.~Thrun, M.~Bennewitz, W.~Burgard, A.~Cremers, F.~Dellaert, D.~Fox, D.~Hahnel,
  C.~Rosenberg, N.~Roy, J.~Schulte, and D.~Schulz, ``Minerva: a
  second-generation museum tour-guide robot,'' in \emph{Proceedings 1999 IEEE
  International Conference on Robotics and Automation (Cat. No.99CH36288C)},
  vol.~3.\hskip 1em plus 0.5em minus 0.4em\relax IEEE, 1999.

\bibitem{shehade_vr_museums_2020}
M.~Shehade and T.~Stylianou-Lambert, ``Virtual reality in museums: Exploring
  the experiences of museum professionals,'' \emph{Applied Sciences}, vol.~10,
  no.~11, 2020.

\bibitem{bekele2018survey}
M.~K. Bekele, R.~Pierdicca, E.~Frontoni, E.~S. Malinverni, and J.~Gain, ``A
  survey of augmented, virtual, and mixed reality for cultural heritage,''
  \emph{Journal on Computing and Cultural Heritage (JOCCH)}, 2018.

\bibitem{reese_learning_2011}
H.~Reese, ``The learning-by-doing principle,'' \emph{Behavioral Development
  Bulletin}, vol.~17, 2011.

\bibitem{hartmann_entertainment_2021}
T.~Hartmann and J.~Fox, ``{Entertainment in Virtual Reality and Beyond: The
  Influence of Embodiment, Co-Location, and Cognitive Distancing on Users’
  Entertainment Experience},'' in \emph{{The Oxford Handbook of Entertainment
  Theory}}.\hskip 1em plus 0.5em minus 0.4em\relax Oxford University Press,
  2021.

\bibitem{javaid_medical_2020}
M.~Javaid and A.~Haleem, ``Virtual reality applications toward medical field,''
  \emph{Clinical Epidemiology and Global Health}, vol.~8, no.~2, 2020.

\bibitem{perez_industrial_2019}
L.~P\'{e}rez, E.~Diez, R.~Usamentiaga, and D.~F. Garc\'{\i}a, ``Industrial
  robot control and operator training using virtual reality interfaces,''
  \emph{Comput. Ind.}, vol. 109, no.~C, 2019.

\bibitem{de_Heuvel_2022}
J.~de~Heuvel, N.~Corral, L.~Bruckschen, and M.~Bennewitz, ``Learning
  personalized human-aware robot navigation using virtual reality
  demonstrations from a user study,'' in \emph{2022 31st IEEE International
  Conference on Robot and Human Interactive Communication (RO-MAN)}.\hskip 1em
  plus 0.5em minus 0.4em\relax IEEE, 2022.

\bibitem{bottega_virtual_2022}
J.~A. Bottega, R.~Steinmetz, A.~H. Kolling, V.~A. Kich, J.~C. de~Jesus, R.~B.
  Grando, and D.~F.~T. Gamarra, ``Virtual reality platform to develop and test
  applications on human-robot social interaction,'' 2022.

\bibitem{stotko_slamcast_2019}
P.~Stotko, S.~Krumpen, M.~B. Hullin, M.~Weinmann, and R.~Klein, ``Slamcast:
  Large-scale, real-time 3d reconstruction and streaming for immersive
  multi-client live telepresence,'' \emph{IEEE Transactions on Visualization
  and Computer Graphics}, vol.~25, no.~5, 2019.

\bibitem{stotko_exploration_2019}
P.~Stotko, S.~Krumpen, M.~Schwarz, C.~Lenz, S.~Behnke, R.~Klein, and
  M.~Weinmann, ``A vr system for immersive teleoperation and live exploration
  with a mobile robot,'' in \emph{2019 IEEE/RSJ International Conference on
  Intelligent Robots and Systems (IROS)}.\hskip 1em plus 0.5em minus
  0.4em\relax IEEE, 2019.

\bibitem{komarac2023understanding}
T.~Komarac and D.~Ozreti\'c~Do\v{s}en, ``Understanding virtual museum visits:
  generation z experiences,'' \emph{Museum Management and Curatorship}, 2023.

\bibitem{andritsou_momap_2018}
G.~Andritsou, A.~Katifori, V.~Kourtis, and Y.~Ioannidis, ``Momap - an
  interactive gamified app for the museum of mineralogy,'' in \emph{2018 10th
  International Conference on Virtual Worlds and Games for Serious Applications
  (VS-Games)}, 2018.

\bibitem{lindemann_collecting_2022}
A.~Lindemann, ``The legacy of a lifetime of collecting: An interactive natural
  history museum exhibit,'' in \emph{10th International Conference on Digital
  and Interactive Arts}, ser. ARTECH 2021.\hskip 1em plus 0.5em minus
  0.4em\relax Association for Computing Machinery, 2022.

\bibitem{cristobal_coral_2020}
F.~R. Cristobal, M.~Dodge, B.~Noll, N.~Rosenberg, J.~Burns, J.~Sanchez,
  D.~Gotshalk, K.~Pascoe, and A.~Runyan, ``Exploration of coral reefs in
  hawai‘i through virtual reality: Hawaiian coral reef museum vr,'' in
  \emph{Practice and Experience in Advanced Research Computing}, ser. PEARC
  '20.\hskip 1em plus 0.5em minus 0.4em\relax Association for Computing
  Machinery, 2020.

\bibitem{beheshti_wires_2017}
E.~Beheshti, D.~Kim, G.~Ecanow, and M.~S. Horn, ``Looking inside the wires:
  Understanding museum visitor learning with an augmented circuit exhibit,'' in
  \emph{Proceedings of the 2017 CHI Conference on Human Factors in Computing
  Systems}, ser. CHI '17.\hskip 1em plus 0.5em minus 0.4em\relax Association
  for Computing Machinery, 2017.

\bibitem{tsita2023virtual}
C.~Tsita, M.~Satratzemi, A.~Pedefoudas, C.~Georgiadis, M.~Zampeti,
  E.~Papavergou, S.~Tsiara, E.~Sismanidou, P.~Kyriakidis, D.~Kehagias,
  \emph{et~al.}, ``A virtual reality museum to reinforce the interpretation of
  contemporary art and increase the educational value of user experience,''
  \emph{Heritage}, 2023.

\bibitem{leow2021analysing}
F.-T. Leow and E.~Ch’ng, ``Analysing narrative engagement with immersive
  environments: designing audience-centric experiences for cultural heritage
  learning,'' \emph{Museum Management and Curatorship}, vol.~36, no.~4, pp.
  342--361, 2021.

\bibitem{garcia_tenochtitlan_2017}
S.~Garcia-Cardona, F.~Tian, and S.~Prakoonwit, ``Tenochtitlan - an interactive
  virtual reality environment that encourages museum exhibit engagement,'' in
  \emph{E-Learning and Games}, F.~Tian, C.~Gatzidis, A.~El~Rhalibi, W.~Tang,
  and F.~Charles, Eds.\hskip 1em plus 0.5em minus 0.4em\relax Springer
  International Publishing, 2017.

\bibitem{schofield_viking_2018}
G.~Schofield, G.~Beale, N.~Beale, M.~Fell, D.~Hadley, J.~Hook, D.~Murphy,
  J.~Richards, and L.~Thresh, ``Viking vr: Designing a virtual reality
  experience for a museum,'' in \emph{Proceedings of the 2018 Designing
  Interactive Systems Conference}, ser. DIS '18.\hskip 1em plus 0.5em minus
  0.4em\relax Association for Computing Machinery, 2018.

\bibitem{li_gibson_2022}
C.~Li, F.~Xia, R.~Mart\'in-Mart\'in, M.~Lingelbach, S.~Srivastava, B.~Shen,
  K.~E. Vainio, C.~Gokmen, G.~Dharan, T.~Jain, A.~Kurenkov, K.~Liu, H.~Gweon,
  J.~Wu, L.~Fei-Fei, and S.~Savarese, ``igibson 2.0: Object-centric simulation
  for robot learning of everyday household tasks,'' in \emph{Proceedings of the
  5th Conference on Robot Learning}.\hskip 1em plus 0.5em minus 0.4em\relax
  PMLR, 2022.

\bibitem{fox1999markov}
D.~Fox, W.~Burgard, and S.~Thrun, ``Markov localization for mobile robots in
  dynamic environments,'' \emph{Journal of artificial intelligence research},
  1999.

\bibitem{thrun_probabilistic_2005}
S.~Thrun, W.~Burgard, and D.~Fox, \emph{Probabilistic robotics}.\hskip 1em plus
  0.5em minus 0.4em\relax MIT Press, 2005.

\bibitem{hart_paths_1968}
P.~E. Hart, N.~J. Nilsson, and B.~Raphael, ``A formal basis for the heuristic
  determination of minimum cost paths,'' \emph{IEEE Transactions on Systems
  Science and Cybernetics}, vol.~4, no.~2, 1968.

\bibitem{fox_dynamic_1997}
D.~Fox, W.~Burgard, and S.~Thrun, ``The dynamic window approach to collision
  avoidance,'' \emph{IEEE Robotics Automation Magazine}, vol.~4, no.~1, 1997.

\bibitem{tucker20museum}
E.~L. Tucker, ``{Museum Studies},'' in \emph{{The Oxford Handbook of
  Qualitative Research}}.\hskip 1em plus 0.5em minus 0.4em\relax Oxford
  University Press, 2020.

\end{thebibliography}

\end{document}